\pdfoutput=1

\documentclass[11pt]{article}

\usepackage[final]{ACL2023}

\usepackage{times}
\usepackage{latexsym}

\usepackage[T1]{fontenc}

\usepackage[utf8]{inputenc}

\usepackage{microtype}

\usepackage{inconsolata}

\usepackage{array,graphicx}
\usepackage{multirow}
\usepackage{booktabs}
\usepackage{pifont}
\usepackage{linguex}

\newcommand*\rot{\rotatebox{90}}
\newcommand*\OK{\ding{51}}
\newcommand*\NO{\ding{55}}

\title{Unifying the Scope of Bridging Anaphora Types in English: \\Bridging Annotations in ARRAU and GUM}

\author{\textbf{Lauren Levine} \and 
\textbf{Amir Zeldes}  \\
  Georgetown University \\
  Department of Linguistics \\
  \texttt{\{lel76, amir.zeldes\}@georgetown.edu}}

\begin{document}
\maketitle
\begin{abstract}

Comparing bridging annotations across coreference resources is difficult, largely due to a lack of standardization across definitions and annotation schemas and narrow coverage of disparate text domains across resources. 
To alleviate domain coverage issues and consolidate schemas, we compare guidelines and use interpretable predictive models to examine the bridging instances annotated in the GUM, GENTLE and ARRAU corpora. Examining these cases, we find that there is a large difference in types of phenomena annotated as bridging.
Beyond theoretical results, we release a harmonized, subcategorized version of the test sets of GUM, GENTLE and the ARRAU Wall Street Journal data to promote meaningful and reliable evaluation of bridging resolution across domains.

\end{abstract}

\section{Introduction}

The term ``bridging'' has been used to describe a broad set of associative coreference phenomena, where the interpretation of an anaphor is in some way dependent on the comprehension of a non-identical antecedent. While considerably less studied than identity coreference, bridging anaphora have been increasingly included in the creation of recent coreference resources, including in recent shared task settings \cite{khosla-etal-2021-codi, yu-etal-2022-codi}. However, bridging annotations are difficult to compare between resources, as corpora frequently differ not only in their text-types and domains, but also in their definitions of bridging as a phenomenon and their annotation schemas for categorizing bridging subtypes \cite{kobayashi-ng-2020-bridging}.\footnote{The same can also be said of definitions of identity coreference, see \citet{Zeldes2022,poesio-etal-2024-universal}; the case of markable span definitions in particular concerns both types of anaphora alike.}  

Due to this difference in both content and schema, it becomes difficult to establish a reliable standard bench-mark for the evaluation of bridging resolution tasks. In this paper we analyze two of the largest available bridging resources for English: GUM \cite{Zeldes2017} and its accompanying test corpus GENTLE \cite{aoyama-etal-2023-gentle}, and the sub-corpora of ARRAU \cite{poesio-artstein-2008-anaphoric, Uryupina2019AnnotatingAB}, with a focus on its largest sub-corpus, ARRAU WSJ, which is composed of Wall Street Journal data. We compare the contents and bridging schemas of these corpora with an eye towards creating more cross resource compatible, high quality bridging data in the future. 

In order to determine significant differences between the corpora, we first find categorical differences in their annotation guidelines and technical formats, and then train predictive models on each corpus, performing error analysis on the cross-corpus prediction results. We also conduct a feature analysis of the predictive models to examine the environmental differences between the occurrences of bridging in the corpora under investigation.
Finally, we provide harmonized test sets for GUM/GENLTE and ARRAU WSJ, providing revised bridging annotations which integrate ARRAU style bridging subtype annotations into GUM and unify the categories for entity type annotations. It is our hope that this effort at harmonization will promote interest in the cross compatibility of bridging resources. 

\section{Background}
\label{sec:background}

\citet{clark-1975-bridging} offers the first theoretical account of bridging as a phenomenon, covering a broad range of discourse inference, including overlap with identity coreference. There have subsequently been various theoretical accounts of bridging which have provided different perspectives \cite{hawkins1978definiteness, asher1998bridging, baumann2012referential}. There have similarly been a number efforts to create annotated resources for bridging, each with its own theoretical understanding of what bridging encompasses as a linguistic phenomenon. 

\citet{kobayashi-ng-2020-bridging} give a survey of currently available bridging datasets, with a focus on English, and list 7 corpora (including 4 sub-corpora of ARRAU), additionally mentioning GUM in passing (the paper predates the release of GENTLE). Table \ref{table:resources} gives an overview of these datasets, and adds and compares some essential properties of their coverage for bridging phenomena. 

\begingroup

\begin{table*} \centering
    \resizebox{\textwidth}{!}{
    \begin{tabular}{@{} cl*{15}c @{}}
        & & \multicolumn{15}{c}{Corpus Characteristics} \\[2ex]
        & & \rot{Domain} & \rot{Docs} & \rot{Tokens} & \rot{Mentions} & \rot{Bridging}
        & \rot{Definite Anaphora} & \rot{Indefinite Anaphora} & \rot{Entity Antecedent} & \rot{Event Antecedent} 
        & \rot{Referential Bridging} & \rot{Lexical Bridging} & \rot{Information Status} 
        & \rot{Subtypes} & \rot{Comparative Anaphora} & \rot{Gold Treebank} \\
        \cmidrule{2-17}
        & ISNotes             & WSJ news & 50 & 40k & 11k & 663 & \OK & \OK & \OK & \OK & \OK & \NO & \OK & \OK & \NO & \OK \\
        & BASHI               & WSJ news & 50 & 58k & 19k & 459 & \OK & \OK & \OK & \OK & \OK & \NO & \NO & \NO & \OK & \OK \\
        & ARRAU RST              & news & 413 & 229k & 72k & 3.7k & \OK & \OK & \OK & \NO & \OK & \OK & \OK & \OK & \OK & \OK \\
        & ARRAU GNOME & \shortstack[c]{medical, art history} & 5 & 21k & 6.5k & 692 & \OK & \OK & \OK & \NO & \OK & \OK & \OK & \OK & \OK & \NO \\
        & ARRAU PEAR & \shortstack[c]{spoken narratives} & 20 & 14k & 4k & 333 & \OK & \OK & \OK & \NO & \OK & \OK & \OK & \OK & \OK & \NO \\
\rot{\rlap{~Corpora}}
        & ARRAU TRAINS & dialogues & 114 & 84k & 17k & 710 & \OK & \OK & \OK & \NO & \OK & \OK & \OK & \OK & \OK &  \NO \\
        & SciCorp & scientific text & 14 & 61k & 9.4k & 1.3k & \OK & \NO & \OK  & \NO & \OK & \NO & \OK & \NO & \OK & \NO \\
        & GUM (V10.1.0) & 16 genres & 235 & 228k & 64k & 1.9k & \OK & \OK & \OK & \OK & \OK & \OK & \OK & \NO & \OK & \OK \\
        & GENTLE (V2.0.0)                & 8 genres & 26 & 18k & 5.6k & 58 & \OK & \OK & \OK & \OK & \OK & \OK & \OK & \NO & \OK & \OK \\
        \cmidrule[1pt]{2-17}
    \end{tabular}
    }
    \caption{Survey of English Bridging Resources}
    \label{table:resources}
\end{table*}

\endgroup

In terms of token count and bridging instances, the news section of ARRAU (ARRAU WSJ; 229k tokens, 3.7k bridging instances) and GUM (228k tokens, 1.9k bridging instances as of version 10) are the largest.\footnote{We include the Reddit subset of GUM for which we obtain the underlying texts using the API \cite{behzad-zeldes-2020-cross}, and use the original markable definitions from GUM, as opposed to the OntoGUM version which is harmonized with OntoNotes definitions \cite{zhu-etal-2021-ontogum}.} While the ARRAU WSJ is only a single genre, GUM includes 16 different genres (academic writing, biographies, courtroom transcripts, essays, fiction, how-to guides, interviews, letters, news, online forum discussions, podcasts, political speeches, spontaneous face to face conversations, textbooks, travel guides, and vlogs), with its extended test corpus GENTLE spanning an additional 8 genres (dictionary entries, live esports commentary, legal documents, medical notes, poetry, mathematical proofs, course syllabuses, and threat letters). Additionally, ARRAU is one of the few corpora which includes subtype annotations for bridging, a feature that GUM and GENTLE lack. The complimentary attributes of these two datasets and their relatively large size of bridging instances makes them prime candidates for comparison and harmonization. As such, in this paper we focus on comparing and unifying between these two bridging schemas, leaving a broader harmonization with other resources for future work. The following section breaks down the categorical differences in these two annotation formalisms.

\section{Categorical Differences}
\label{sec:differences}

Some of the most substantial differences between the datasets come from their theoretical underpinnings and technical infrastructure. In ARRAU, bridging is considered to be a type of ``anaphoric reference which links the object being referred to by the markable to an already established discourse entity ... via a semantic relation other than coreference'' \cite{arrau3-annotation-manual}. This focus on semantic relations takes a more lexically grounded approach, laying out specific types of semantic relations to be marked as bridging, including \textit{part-of} and \textit{set} relations. GUM/GENTLE, by contrast, approach bridging from the perspective of information status, broadly laying out bridging as any newly introduced entity which is in some way underspecified, but whose identity is interpretable/inferable thanks to a non-identical antecedent entity \cite{gum-wiki}. The main structural differences that emerge from a comparison of the datasets' guidelines and annotations are laid out below:

\paragraph{Previously mentioned anaphors} While in GUM/GENTLE the entity of a bridging anaphor must be mentioned for the first time after its antecedent has already been introduced, ARRAU considers bridging to apply even if the entity in question has already been introduced, as in \ref{ex:prev-mention}.

\ex. \textit{Could I move .. $[$engine E two$]_i$ .. there should be $[$one engine$]_j$ at Corning .. $[$engine E two$]_i$ is there}\label{ex:prev-mention}

\noindent In this example, the second mention of \textit{engine E two} is annotated in ARRAU as a bridging anaphor to \textit{one engine}; however the entity \textit{engine E two} was already introduced into the discourse earlier, and is also annotated as a coref antecedent to the bridging anaphor. Such examples are prohibited in GUM/GENTLE where bridging anaphora can only occur with non-given (i.e.~non-aforementioned) mentions.

\paragraph{Split bridging antecedents} ARRAU allows bridging from one anaphor to multiple antecedents as in \ref{ex:multiple-ante}.
    
\ex. \textit{FOREIGN PRIME RATES: $[$Canada 13.50\%$]_i$; $[$Germany 8.50\%$]_j$; $[$Japan 4.875\%$]_k$; $[$Switzerland 8.50\%$]_l$; $[$Britain 15\%$]_m$ .. lending practices vary widely by $[$location$]_n$}\label{ex:multiple-ante}

\noindent Here the mention \textit{location} is taken as an anaphor bridging to all pairs of country and prime rate. While it is true that it refers to the location of a loan, we question whether the antecedent is truly split: if the location is the countries, then this is split antecedent coreference (and not bridging); if it is inferrable from the existence of a rate, which has a jurisdiction location which is conceptually distinct from the country, then the antecedent should simply be \textit{FOREIGN PRIME RATES}. Regardless, such split bridging antecedents are categorically excluded in GUM.

\paragraph{Discontinuous mention spans} In ARRAU entity spans are allowed to be discontinuous spans of tokens, while GUM's representation format does not support discontinuous mentions. This allows for a more appropriate handling of spans such as ``\textit{$[$Mr.$]_{i1}$ and $[$Mrs. $[$Smith$]_{i2}]_j$}'', where the indices \textit{i1} and \textit{i2} indicate the two parts of the 
discontinuous mention ``Mr. Smith''. In GUM/GENTLE, continuous spans represent both mentions, spuriously including `Mrs.' in the first mention: ``\textit{$[$Mr. and $[$Mrs. Smith$]_{j}]_i$}''.

\paragraph{Entity types} ARRAU allows for a coreference cluster to contain multiple entity types amongst its members (e.g.~\textsc{organization} and \textsc{location} for a country), while GUM requires that all members of a coreference cluster have the same entity type. Additionally and unlike GUM, ARRAU does not assign an entity type for coordinate entity mentions, for instance in \ref{ex:coordinate}: 

\ex. \textit{$[$ $[$wildlife$]_{\textsc{animate}}$ and $[$the fishing industry$]_{\textsc{abstract}}$ $]_{\textsc{none}}$}\label{ex:coordinate}

Even though the entities within the coordination have types (\textsc{animate} and \textsc{abstract}), the entity type of the coordinate phrase is \textsc{none}. In GUM,  coordinate entities only receive a shared markable if they are also referred to in aggregate elsewhere in the text. In the case of a mixed type coordinate markable, the coordinate phrase and related aggregate mention will both be labeled with the entity type \textsc{abstract}. 

The inventory of possible entity types in ARRAU and GUM also have some minor differences. For our purposes, we create a unified set of entity categories and collapse the inventories of the individual resources as shown in Table \ref{table:entity_types}.

\begin{table}[]
\centering
\resizebox{\columnwidth}{!}{%
\begin{tabular}{lll}
\hline
\multirow{2}{*}{\textbf{Unified Types}} & \multicolumn{2}{l}{\textbf{Original Entity Types}} \\ \cline{2-3} 
 & \multicolumn{1}{l}{\textbf{GUM}} & \multicolumn{1}{l}{\textbf{ARRAU}} \\ \hline
\textsc{person} & \multicolumn{1}{l}{\textsc{person}} & \textsc{person} \\ 
\textsc{place} & \multicolumn{1}{l}{\textsc{place}} & \textsc{space} \\ 
\textsc{organization} & \multicolumn{1}{l}{\textsc{organization}} & \textsc{organization} \\ 
\textsc{concrete} & \multicolumn{1}{l}{\textsc{object, plant}} & \textsc{concrete} \\ 
\textsc{event} & \multicolumn{1}{l}{\textsc{event}} & \textsc{plan} \\ 
\textsc{time} & \multicolumn{1}{l}{\textsc{time}} & \textsc{time} \\ 
\textsc{substance} & \multicolumn{1}{l}{\textsc{substance}} & \textsc{substance, medicine} \\ 
\textsc{animate} & \multicolumn{1}{l}{\textsc{animal}} & \textsc{animate} \\ 
\textsc{abstract} & \multicolumn{1}{l}{\textsc{abstract}} & \begin{tabular}[c]{@{}l@{}}\textsc{abstract, undersp-onto,} \\ \textsc{disease, numerical, none}\end{tabular} \\ \hline
\end{tabular}%
}
\caption{Unified entity types between the GUM and ARRAU schemas}
\label{table:entity_types}
\end{table}

\paragraph{Bridging subtypes} While GUM does not attempt to subcategorize different varieties of bridging, ARRAU WSJ has an inventory of 9 subtype labels which can be applied to bridging annotations. Table \ref{table:subtype_desc} lists the different bridging subtype labels used in ARRAU, along with a brief explanation for each. Further explanation can be found in ARRAU's annotation guidelines \cite{arrau3-annotation-manual}.

\paragraph{} Reflecting on these categorical differences, we favor GUM's more structurally restrictive approach, which is based on the information status of mentions, since it links the phenomenon to the cognitive act of bridging as a form of information fetching: if an anaphor requires back reference to resolve but has not been mentioned before, then bridging has occurred. Semantic criteria, by contrast, are less easy to apply, since we find many NPs whose extensions can be considered to form set-member relations but are not annotated as bridging in ARRAU, from `people' to any person mentioned in a text to ontological categories (e.g. `time' in general vs. specific times), and we would like to exclude such cases on principled grounds. 

Additionally, we like that GUM's approach casts a wider scope in terms of possible bridging varieties because it does not depend on a finite set of pre-defined semantic relations to identify instances of bridging, whereas ARRAU appears to depend on such pre-defined relations. However, we do find the additional granularity of the bridging subtypes in ARRAU to be desirable. As such, we advocate for a less restrictive approach to identifying bridging relations, as in GUM, which can then be categorized with more granular subtype relations, such as those used in ARRAU. This view is reflected in our test set harmonization effort detailed in Section \ref{sec:test_sets}.

\begin{table}[]
\centering
\resizebox{\columnwidth}{!}{%
\begin{tabular}{ll}
\hline
\multicolumn{1}{c}{\textbf{Subtype}} & \multicolumn{1}{c}{\textbf{Description}} \\ \hline
\textsc{poss} & anaphor is a part/attribute of the antecedent \\
\textsc{poss-inv} & antecedent is a part/attribute of the anaphor \\
\textsc{element} & anaphor is an element of the antecedent set \\
\textsc{element-inv} & antecedent is an element of the anaphor set \\
\textsc{subset} & anaphor is a subset of the antecedent set \\
\textsc{subset-inv} & antecedent is subset of the anaphor set \\
\textsc{other} & anaphor marked with "other" \\
\textsc{other-inv} & antecedent marked with "other" \\
\textsc{undersp-rel} & sense anaphora, situational reference \\
(unmarked) & no subtype annotation was provided \\ \hline
\end{tabular}%
}
\caption{Bridging subtype labels used in ARRAU}
\label{table:subtype_desc}
\end{table}

\section{Predictive Models}
\label{sec:models}

Although the differences outlined in Section \ref{sec:differences} are the most striking, and responsible for the largest discrepancies in bridging frequency and included subtypes, a long tail of less obvious differences distinguishes much of the data in the different corpora. In order to identify such subtle differences, we train and test bridging mention classifiers across corpora, starting with gold standard mention spans and trying to answer the question: which bridging instances in one corpus would not be considered ones in the other, and which instances does one corpus miss, which another might include? Given the sparseness of the data, exacerbated by the need to fine-tune a separate model on each dataset, we use statistical machine learning models and attempt to extract consistent features for mention spans from each corpus. 

\subsection{Data}
\label{sec:data}

Based on the categorical differences outlined in Section \ref{sec:differences}, we remove cases annotated as bridging in ARRAU WSJ which we believe are structurally ineligible to be instances of bridging in GUM, and which would otherwise compromise compatibility between datasets. From the ARRAU WSJ data, we remove 297 cases of bridging with multiple antecedents, and 957 instances where the bridging anaphor already had an identity coreference antecedent. Although this reduction of over 1,200 cases necessarily loses information, we observe that the much tighter information structural definition of bridging leads to more consistency in example types, and note that this already accounts for most of the difference in bridging prevalence between the datasets, leaving ARRAU WSJ with about 12 bridging instances per 1K tokens (and not the unfiltered 18.2), compared to GUM's 8.3 instances per 1K tokens.

We also found that there are 864 instances of discontinuous entity span instances which we include, but for consistency treat as continuous, emulating the behavior in GUM/GENTLE.\footnote{Though out of scope for this paper, we believe that discontinuous mentions are the more accurate analysis and could be introduced into the GUM data, possibly using the gold syntax trees.} A count of the remaining bridging instances in each dataset is shown in Table \ref{table:data_bridging_count}. The harmonized test sets presented in Section \ref{sec:test_sets} are composed of this reduced set of bridging annotations.

We compose separate train and test datasets for GUM/GENTLE and ARRAU, so we may analyze their bridging environments separately. The training data for the GUM classifier contains documents from GUM's given train and dev partitions, while the test data contains documents from GUM's test partition and GENTLE, to test texts that are out-of-domain in both datasets. The training data for the ARRAU classifier contains documents from ARRAU WSJ's given train and dev partitions, and test contains documents from ARRAU WSJ's given test partition. 

In order to train and evaluate our predictive models for bridging, we formulate the task as a binary judgement: given a pair of mentions and their accompanying linguistic features, predict whether the pair is an instance of bridging. We first extract all mention instances from the GUM/GENTLE and ARRAU data, and then enumerate all possible pairs, tracking whether they are instances of bridging, identity coreference, or non-coreference pairs. For each extracted mention, we track the entity type (which has been collapsed to be compatible between the two corpora, as in Table \ref{table:entity_types}), information status, definiteness, phrase length, distance, and the following attributes of the syntactic head of each entity: dependency relation (deprel), part of speech (xpos), lemma, and number (plural vs. singular). To obtain dependency relations for ARRAU WSJ, we convert the gold constituent trees to dependencies using CoreNLP \cite{ManningEtAl2014}.

Due to the relative scarcity of bridging instances, we construct each train and test set with balanced classes of bridging, identity coreference, and non-coreference pairs. We first take all of the bridging pairs in the documents from each section of the data partitions (counts shown in Table \ref{table:data_bridging_count}), and then we take a random selection of an equal quantity for identity coreference and non-coreference pairs. In order to have reasonable candidate pairs in this selection, pronoun anaphora are excluded, as they are almost always instances of identity coreference, and non-bridging cases are limited to anaphor-antecedent pairs that are within the maximum distance of an attested bridging pair.

\begin{table}[]
\centering
\resizebox{\columnwidth}{!}{%
\begin{tabular}{ccc}
\hline
 & \textbf{GUM/GENTLE} & \textbf{ARRAU WSJ} \\ \hline
\textbf{Train} & 1611 & 779 \\
\textbf{Test} & 280 & 176 \\ \hline
\end{tabular}%
}
\caption{Counts of bridging instances in classifier training and evaluation data}
\label{table:data_bridging_count}
\end{table}

\subsection{Models}
\label{sec:sub-models}

Using the data partitions for each corpus outlined in Section \ref{sec:data}, we train two XGBoost classifiers\footnote{\url{https://xgboost.readthedocs.io/en/latest/index.html}}: one trained on the train data from GUM, and the other trained on the train data from ARRAU WSJ. Each of these classifiers was trained and optimized with a grid search with 5 fold cross validation on the training data, using a subset of the linguistic features extracted during the creation of the mention pair data partitions. For both the antecedent and the anaphor of the mention pair, features included entity type, definiteness, phrase length, and syntactic dependency relation, part of speech, number, and lemma of the mention's syntactic head. Additionally, the information status of the antecedent and antecedent-anaphor distance were included as features. 

Table \ref{table:classifer_results} shows the performance of each classifier on its own test data and the test data of the other corpus, along with the performance of a random baseline on both test sets (averaged from 5 runs). The random baseline has a 33\% chance of predicting an antecedent-anaphor pair as bridging, reflecting the balanced classes of bridging, identity coreference, non-coreference pairs in the test sets. Even with the classes balanced in the train and test data, we see that each classifier's performance on predicting the positive class of bridging  on their own test data is relatively low, with the GUM classifier giving an F-score of 0.71, and the ARRAU classifier giving an F-score of 0.67. Still, we see that these both substantially outperform the random baseline, which gives an F-score of 0.32 on GUM/GENTLE test and an F-score of 0.33 on ARRAU WSJ test.

The performance of each classifier on the test data of the other corpus is lower than on its own, with the ARRAU classifier giving an F-score of 0.56 on the GUM/GENTLE eval data and the GUM classifier giving an F-score of just 0.22 (worse than the random baseline), on the ARRAU WSJ eval data. Given the substantial differences in the approach to bridging annotations between the two corpora, performance degradation on cross-corpus prediction is expected. However, it is worth noting that the performance degradation is steeper for the GUM classifier than the ARRAU classifier, to the point where the GUM classifier performs worse than random chance. This suggests that ARRAU may have more varieties of bridging not seen in the GUM data than vice versa. In order to investigate the differences in bridging varieties in the two corpora more closely, we conduct an analysis of the feature importance of the predictive models in Section \ref{sec:feature-analysis}, and an error analysis of the cross-corpus prediction results in Section \ref{sec:error-analysis}.

\begin{table}[]
    \centering
    \resizebox{\columnwidth}{!}{%
    \begin{tabular}{lllll}
    \hline
    \textbf{Classifiers} & \textbf{Eval Data} & \textbf{P} & \textbf{R} & \textbf{F} \\ \hline
    \multirow{2}{*}{GUM} & GUM/GENTLE & 0.74 & 0.68 & 0.71 \\
     & ARRAU WSJ & 0.41 & 0.15 & 0.22 \\ \hline
    \multirow{2}{*}{ARRAU WSJ} 
    & GUM/GENTLE & 0.57 & 0.55 & 0.56 \\
    & ARRAU WSJ & 0.66 & 0.69 & 0.67 \\ \hline
     \multirow{2}{*}{Random Baseline} 
     & GUM/GENTLE & 0.32 & 0.32 & 0.32 \\
     & ARRAU WSJ & 0.33 & 0.34 & 0.33 \\ \hline
    \end{tabular}%
    }
    \caption{XGBoost classifier performance and random baseline on predicting the positive class of bridging cases}
    \label{table:classifer_results}
\end{table}

\begin{figure}
  \centering
  \includegraphics[width=1\linewidth]{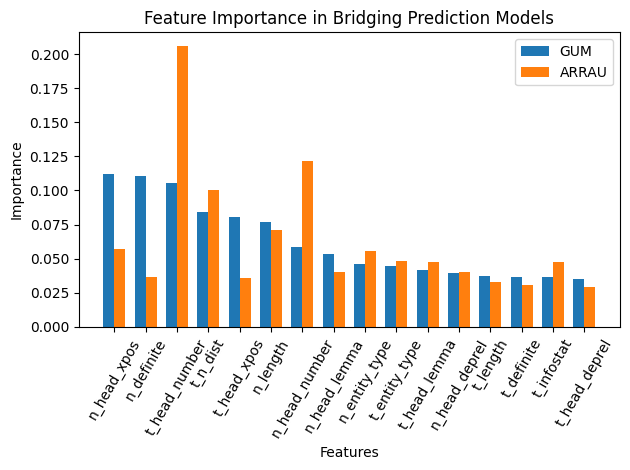}
  \caption{Feature importance of XGBoost classifiers trained on GUM and ARRAU WSJ}
  \label{fig:feat_import}
\end{figure}

\section{Feature Analysis}
\label{sec:feature-analysis}

Feature importance is an indication of the relative contribution of a particular feature for the decision of a model. By examining the feature importances of our predictive models from Section \ref{sec:sub-models}, we can gain insight into which linguistic features are characteristic of the varieties of bridging captured in each of the corpora used for model training. The feature importance results of the GUM classifier and the ARRAU classifier are shown in Figure \ref{fig:feat_import}. Importance is measured using XGBoost's importance type ``gain'', which indicates the average contribution of the corresponding feature over the trees in the model based on the purity metric Gini. For comparison, we include the feature importance of the models using Mean Decrease in Accuracy (MDA) as a metric in Appendix \ref{sec:appendix}. Looking at Figure \ref{fig:feat_import}, we see that the part of speech of the anaphor and the definiteness of the anaphor are the features of most import for the GUM classifier, while the number (plural vs. singular) of the antecedent and the anaphor are the most important features for the ARRAU classifier. Number is perhaps such an important feature in the ARRAU classifier due to the focus on capturing the pre-defined set-element and set-subset relations as instances of bridging.

As definiteness of a newly introduced entity is a strong signal of some form of referential bridging, it is logical to see definitenss of the anaphor as an important feature for the GUM classifier. It has a much lower relative importance for the ARRAU classifier, possibly because ARRAU is not limited to newly introduced entities as candidates for bridging, focusing more on semantic part-whole or subset relations. In Table \ref{table:residuals}, we show the chi-square residuals for definiteness of the anaphor as an indication of an entity pair being an instance of bridging. We see that in both GUM/GENTLE and ARRAU WSJ, a definite anaphor is a positive indicator for bridging and an indefinite anaphor is negative indicator for bridging. However, the magnitude of the residuals for the GUM/GENTLE data is notably larger than those of the ARRAU WSJ. This confirms that definiteness of the anaphor is a stronger indication of whether something is bridging in the GUM/GENTLE data than in the ARRAU WSJ data.

\begin{table}[]
\centering
\resizebox{\columnwidth}{!}{%
\begin{tabular}{llclllc}
\cline{1-3} \cline{5-7}
 & Bridge & \multicolumn{1}{l}{Non-bridge} &  &  & Bridge & \multicolumn{1}{l}{Non-bridge} \\ \cline{1-3} \cline{5-7} 
Def & \multicolumn{1}{c}{13.3} & -9.4 &  & Def & \multicolumn{1}{c}{0.7} & -0.5 \\
Ind & \multicolumn{1}{c}{-7.4} & 5.3 &  & Ind & \multicolumn{1}{c}{-0.3} & 0.2 \\ \cline{1-3} \cline{5-7} 
 & \multicolumn{2}{l}{GUM/GENLTE} &  &  & \multicolumn{2}{l}{ARRAU WSJ} \\ \cline{1-3} \cline{5-7} 
\end{tabular}%
}
\caption{Chi-square residuals for definiteness of the anaphor (definite vs. indefinite) being an indicator of bridging in GUM/GENTLE and ARRAU WSJ}
    \label{table:residuals}
\end{table}

\begin{figure}[h!tb]
  \centering
  \includegraphics[width=1\linewidth]{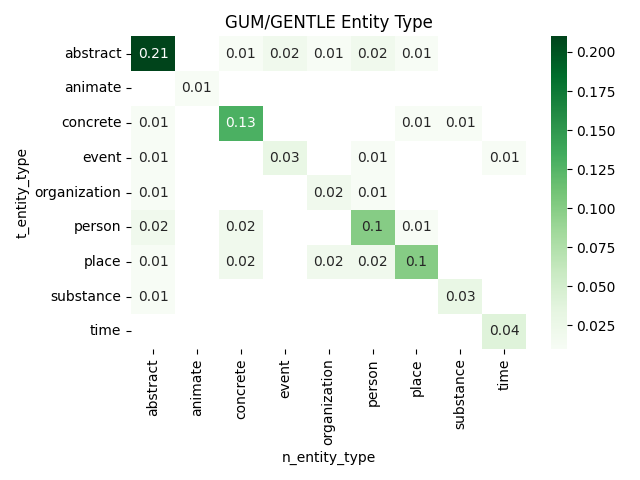}
  \caption{Distribution for antecedent-anaphor entity type combinations for GUM/GENTLE (only combinations with a proportion of 1\% or higher are visualized)}
  \label{fig:gum_entities}
\end{figure}

\begin{figure}[h!tb]
  \centering
  \includegraphics[width=1\linewidth]{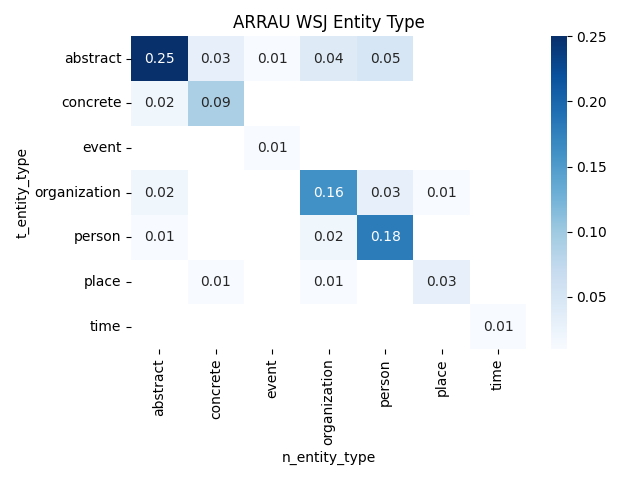}
  \caption{Distribution for antecedent-anaphor entity type combinations for ARRAU WSJ (only combinations with a proportion of 1\% or higher are visualized)}
  \label{fig:arrau_entities}
\end{figure}

The entity types of the antecedent and the anaphor of a bridging instance are a set of categorical features with similar feature importance in the two classifiers. To investigate these features jointly, in Figures \ref{fig:gum_entities} and \ref{fig:arrau_entities} we provide heatmaps of the distributions of antecedent-anaphor entity type combinations in bridging instances in each dataset. We can see that in both datasets it is common for bridging to occur to and from entities of the same type. We also see that bridging to and from \textsc{abstract} entities is common in both datasets. However, we also see that in GUM bridging instances of entity type \textsc{place-place} are one of the more common combinations (10\% compared to 3\% in ARRAU), while in ARRAU WSJ bridging instances of entity type \textsc{organization-organization} are of higher frequency (16\% compared to 2\% in GUM). Such differences in distribution indicate that there is variation between the resources, either due to differences in corpus content or differences in bridging varieties annotated. In the following section, we investigate some concrete examples within the test sets of each corpus.

\section{Cross-Corpus Error Analysis}
\label{sec:error-analysis}

As we can see from observing the prediction scores of the models in Table \ref{table:classifer_results}, the GUM and ARRAU classifiers have moderate success in predicting instances of bridging in their own test sets, but see performance degrade when applied to the test set of the other corpus. This tells us that the classifiers have learned some characteristic features of their respective training datasets. 
Using the decision probabilities outputted from the classifiers as a confidence measure, we can look at examples which the classifiers are most confident about but predicted incorrectly in order to look for characteristic differences between the bridging instances included in each dataset. Below we look at some of the most confident mistakes of the classifiers on the test data of the opposite corpus.

Memorization of specific noun pairs, such as ``house''--``door'' or ``country''--``capital'' is an important tool in predicting bridging relations in unseen data, which is unsurprisingly more effective for common nouns. It therefore comes as no surprise that many of the GUM classifier's errors on the ARRAU test set seem to stem from out of vocabulary (OOV) items, due to the large number of named entities unique to the WSJ domain. In fact, 10.1\% of RST-DT tokens are proper nouns, compared to just 5.8\% in GUM. This creates a large amount of noise in the error pool, which made it difficult to find example cases that exposed characteristic differences between the datasets. Additionally, it is worth noting that the low performance of the GUM classifier on the ARRAU data (worse than chance) brings into question the utility of analyzing individual examples of incorrect predictions. However, mistakes of the ARRAU classifier on the GUM test set highlighted several common bridging situations which are present in GUM due to its genre diversity, but are missing from ARRAU WSJ, which only has news data. Out of the 280 samples in the GUM/GENTLE test set, there are 18 instances which the ARRAU classifier gives a <10\% probability of being instances of bridging even though they actually are instances of bridging. 

For example, the ARRAU classifier gives a <1\% probability that the boxed entities in \ref{ex:reddit-you-i} are an example of bridging, though it is annotated as such in GUM.

\ex. \textit{Escape The Room Employees, what is the weirdest thing $[$you$]$'ve seen someone do in one of the rooms?\\\\OH WAIT $[$I$]$ THOUGHT OF ANOTHER ONE}\label{ex:reddit-you-i}

GUM allows bridging in cases where multiple addressees are later referenced individually, as in the case above where a question to multiple addresses on an online discussion forum is answered by an individual's post. 

Another genre specific example comes from person-to-heading bridging instances in GUM, which are common in the biography genre of the corpus. For instance, the ARRAU classifier gives example \ref{ex:bio-heading} from a biography text in GUM only a 5\% probability of being bridging.

\ex. \textit{Jens Otto Harry Jespersen...was $[$a Danish linguist who specialized in the grammar of the English language$]$\\\\$[$Early life$]$}\label{ex:bio-heading}

The above example is an instance of bridging from a person (a Danish linguist), to a heading which one infers is a reference to that individual due to the expected structure of a biographical text (early life = the early life of the Danish linguist under discussion).

Similarly, in GUM's academic genre, bridging instances to and from various captions and citations are a direct result of the graphical organization of the text type. For example, the ARRAU classifier gives the following example of bridging a probability of only 7\%:

\ex. \textit{$[$Figure 2.2$]$\\\\A pre-1982 copper penny ( $[$left$]$ ) contains approximately 3 ×× 10 22 copper atoms...}

In the example above, a figure citation is bridging to an entity within the caption of the figure, which references an internal part of the figure itself (the left part of the figure). While these sort of part-whole relations are a very common form of bridging, the application to graphical references is a genre specific phenomenon that one would not necessarily observe in any given corpus of bridging.

From the examples above, we can see that the genre of a text can play a big role in the types of bridging that will be present. As the ARRAU corpus does not include online forum discussions, biographies, and academic texts, it is not able to represent the varieties of bridging which are characteristic to these genres. This gap in coverage contributes to the motivation to have a larger number of comparable bridging resources from a diverse set of domains.

\section{Harmonized Test Sets}
\label{sec:test_sets}

In order to promote the comparability of cross-corpus evaluation results for bridging resolution systems (see \citealt{hou-2020-bridging,kobayashi-ng-2021-bridging}), we present harmonized test sets for GUM/GENTLE and ARRAU WSJ\footnote{\url{https://github.com/lauren-lizzy-levine/bridging_test_sets}}. For these revised test sets, we harmonize on the following three points: categorical differences regarding the scope of bridging, categories for entity types, and bridging subtype annotations.

As discussed in Section \ref{sec:differences}, there are several categorical differences in the definitions of what counts as bridging in GUM and ARRAU. For the purpose of unifying the scope of bridging between these two corpora, we favor GUM's more structurally restrictive approach. As such, we remove cases of bridging with multiple antecedents and cases where the bridging anaphor has an identity coreference antecedent from the ARRAU WSJ test set. This leaves us with 176 instances of bridging in the revised ARRAU WSJ test set.

As ARRAU and GUM have relatively similar entity type categories in their original schemas for entity annotations, we are able to combine them into a single condensed set as shown in Table \ref{table:entity_types}. While entity types are not integral to the comparability of bridging annotations, they are a relevant feature for analysis, so we choose to include them in this harmonization effort. The distribution of bridging anaphor entity types in each of the test sets is shown in Table \ref{table:test_entities}.

As noted in Section \ref{sec:differences}, ARRAU has a schema for categorizing subtypes of bridging (shown in Table \ref{table:subtype_desc}), while GUM does not make any attempt to differentiate sub-varieties of bridging. As such, we harmonize the bridging annotations by manually annotating the GUM/GENTLE test set with ARRAU style bridging subtypes. This annotation was completed by the authors of this paper. While manually annotating the bridging instances in the GUM/GENTLE test set with ARRAU style bridging subtypes, 8 instances of bridging were thrown out as annotation errors, leaving 272 instances of bridging in the GUM/GENTLE test set. 

The distribution of bridging subtypes in each of the test sets is shown in Table \ref{table:test_subtypes}. In both test sets, \textsc{element} and \textsc{subset} are common bridging subtypes, but we see that GUM/GENTLE have larger proportions of the \textsc{poss} and \textsc{undersp-rel} categories. This suggests some difference in the bridging varieties between the two corpora, but is likely also partially explained by the large portion of bridging instances in the ARRAU test set which did not receive a bridging subtype annotation (36\%). We leave experimental evaluation of systems on these harmonized test sets for future work and hope that their availability will promote more meaningfully comparable research on bridging resolution across a range of text types.

\begin{table}[]
\centering
\resizebox{\columnwidth}{!}{%
\begin{tabular}{lll}
\hline
\multicolumn{1}{l}{\textbf{Entity Types}} & \multicolumn{1}{l}{\textbf{GUM/GENTLE}} & \multicolumn{1}{l}{\textbf{ARRAU WSJ}} \\ \hline
\textsc{person} & 40 (15\%) & 41 (23\%) \\
\textsc{place} & 45 (17\%) & 11 (6\%) \\
\textsc{organization} & 18 (7\%) & 55 (31\%) \\
\textsc{concrete} & 57 (21\%) & 22 (13\%) \\
\textsc{event} & 20 (7\%) & 2 (1\%) \\
\textsc{time} & 15 (6\%) & 2 (1\%) \\
\textsc{substance} & 6 (2\%) & 0 \\
\textsc{animate} & 6 (2\%) & 0 \\
\textsc{abstract} & 63 (23\%) & 43 (24\%) \\ \hline
\end{tabular}%
}
\caption{Distribution of bridging anaphor entity types in harmonized GUM/GENTLE test and ARRAU WSJ test.}
\label{table:test_entities}
\end{table}

\begin{table}[]
\centering
\resizebox{\columnwidth}{!}{%
\begin{tabular}{lll}
\hline
\multicolumn{1}{l}{\textbf{Bridging Subtype}} & \multicolumn{1}{l}{\textbf{GUM/GENTLE}} & \multicolumn{1}{l}{\textbf{ARRAU WSJ}} \\ \hline
\textsc{poss} & 78 (29\%) & 9 (5\%) \\
\textsc{poss-inv} & 4 (1\%) & 2 (1\%) \\
\textsc{element} & 81 (30\%) & 49 (28\%) \\
\textsc{element-inv} & 17 (6\%) & 5 (3\%) \\
\textsc{subset} & 21 (8\%) & 33 (19\%) \\
\textsc{subset-inv} & 5 (2\%) & 9 (5\%) \\
\textsc{other} & 11 (4\%) & 3 (2\%) \\
\textsc{other-inv} & 6 (2\%) & 1 (<1\%) \\
\textsc{undersp-rel} & 49 (18\%) & 1 (<1\%) \\
(unmarked) & 0 & 64 (36\%) \\ \hline
\end{tabular}%
}
\caption{Distribution of bridging subtypes in harmonized GUM/GENTLE test and ARRAU WSJ test.}
\label{table:test_subtypes}
\end{table}

\section{Conclusion}
\label{sec:conclusion}

In this paper, we compared the bridging annotations from two of the largest English language corpora with such annotations: ARRAU and GUM. We examined the categorical differences between the scope of their definitions for bridging, and the subtypes annotated within each corpus. We also used predictive models to analyze the linguistic environments and finding examples of interesting differences between the bridging varieties included in each corpus. These differences stem from not only the different genre composition of the corpora, but also the approach towards bridging taken by each corpus. This finding encourages the creation of more genre diverse resources for bridging that are readily comparable with existing resources for bridging. To this end, we have also provided harmonized versions of the GUM/GENTLE test set and the ARRAU WSJ test set, which include unified entity types and ARRAU style bridging subtype annotations added to GUM/GENTLE test. We intend for these harmonized test sets to be the beginning of a larger effort to create a more unified, cross compatible ecosystem of bridging resources for linguistic research and work on automatic bridging resolution.

\section*{Limitations}

This project is the beginning of an effort to create a more uniform and cross-compatible ecosystem of bridging resources, so it naturally leaves much for future work. In this work, we only examine two of the existing English resources for bridging, and we do not consider the annotation schemas of resources for other languages (such as for German \cite{grishina-2016-experiments, Eckart2012}). Subsequent work will require a broader consideration of the various phenomena captured under the label of bridging in various resources and their accompanying categorization schemas.


\bibliography{anthology,custom}
\bibliographystyle{acl_natbib}

\appendix

\section{Mean Decrease Accuracy Feature Importance}
\label{sec:appendix}

\begin{figure}
  \centering
  \includegraphics[width=1\linewidth]{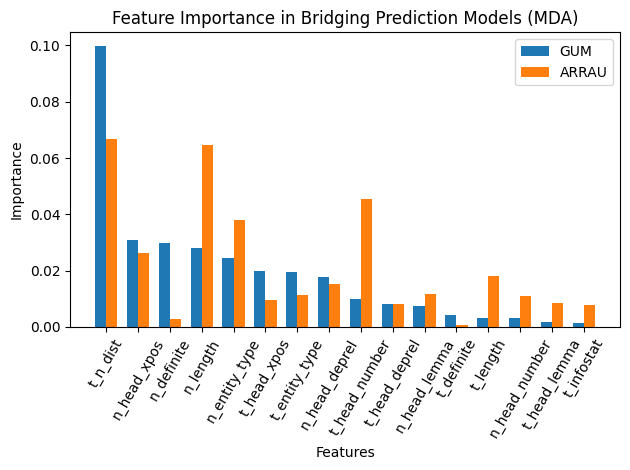}
  \caption{Feature importance of XGBoost classifiers trained on GUM and ARRAU WSJ for Mean Decrease Accuracy (MDA)}
  \label{fig:mda_feat_import}
\end{figure}

For the sake of comparison with our original feature importance analysis shown in Figure \ref{fig:feat_import}, we include the feature importance of the models using Mean Decrease in Accuracy (MDA) as a metric in Figure \ref{fig:mda_feat_import}. Comparing the two figures, we see that the feature importance results are somewhat different between the two metrics. Using MDA as a metric, in both the GUM classifier and the ARRAU classifier, the feature with the most importance is the distance between the antecedent and the anaphor (t\_a\_dist), which was not the case using the Gini based metric. However, by analyzing the feature importance for our models using two different metrics and examining their overlap, we can also see which features are consistently important for each model. The part of speech of the anaphor head (n\_head\_xpos) and the definiteness of the anaphor (n\_definite) remain in the top three most important features for the GUM classifier when using MDA as a metric. Additionally, the number (plural vs. singular) of the antecedent (t\_head\_number) remains in the three most important features for the ARRAU classifier when using MDA as a metric.

\end{document}